\pdfoutput=1

\documentclass[11pt]{article}

\usepackage[final]{acl}
\usepackage{cuted}
\usepackage{caption}  

\usepackage{amsmath} 
\usepackage{booktabs} 
\usepackage{times}
\usepackage{latexsym}
\usepackage{subcaption}
\usepackage{colortbl}
\usepackage{booktabs}
\usepackage{array}
\usepackage{multirow}
\usepackage{enumerate}
\usepackage{adjustbox} 

\usepackage{xcolor}

\usepackage[T1]{fontenc}

\usepackage[utf8]{inputenc}

\usepackage{microtype}

\usepackage{inconsolata}
\usepackage{listings}
\usepackage{color}
\usepackage{algorithm} 
\usepackage{algpseudocode} 
\usepackage{graphicx} 
\usepackage{float}    
\usepackage{pifont}

\usepackage{times}
\usepackage{microtype}
\usepackage{epsfig}
\usepackage{float}
\usepackage{placeins}
\usepackage{color, colortbl}
\usepackage{stfloats}
\usepackage{enumitem}
\usepackage{tabularx}
\usepackage{xstring}
\usepackage{multirow}
\usepackage{xspace}
\usepackage{hyperref}
\usepackage{url}
\usepackage{xcolor}
\usepackage[hang,flushmargin]{footmisc}

\definecolor{codegreen}{rgb}{0,0.6,0}
\definecolor{codegray}{rgb}{0.5,0.5,0.5}
\definecolor{codepurple}{rgb}{0.58,0,0.82}
\definecolor{backcolour}{rgb}{0.95,0.95,0.92}
\definecolor{mypurple}{RGB}{200,192,248}
\definecolor{mypurpledeep}{RGB}{142,126,240}
\definecolor{mygreen}{RGB}{117,170,156}
\definecolor{myyellow}{RGB}{255,192,0}
\definecolor{myblue}{RGB}{57,143,255}
\definecolor{mygrey}{RGB}{231,230,230}
\definecolor{codey}{RGB}{220,220,170}
\definecolor{coder}{RGB}{206,145,120}
\definecolor{codeb}{RGB}{156,220,254}
\definecolor{codenum}{RGB}{204,204,204}
\definecolor{headerblue}{RGB}{224,235,245}

\usepackage{multicol} 

\usepackage[most]{tcolorbox} 

\newtcolorbox{analysisbox}[1][]{
    enhanced jigsaw,
    colback=white,
    colframe=blue!75!black,
    fonttitle=\bfseries,
    boxsep=5pt,
    left=5pt,
    right=5pt,
    top=5pt,
    bottom=5pt,
    title=#1,
}

\newtcolorbox{analysisboxcode}[1][]{
    enhanced jigsaw,
    colback=white,
    colframe=yellow!75!black,
    fonttitle=\bfseries,
    boxsep=5pt,
    left=5pt,
    right=5pt,
    top=5pt,
    bottom=5pt,
    title=#1,
}

\lstdefinestyle{mystyle}{
    backgroundcolor=\color{backcolour},   
    commentstyle=\color{codegreen},
    keywordstyle=\color{magenta},
    numberstyle=\tiny\color{codegray},
    stringstyle=\color{codepurple},
    basicstyle=\footnotesize,
    breakatwhitespace=false,         
    breaklines=true,                 
    captionpos=b,                    
    keepspaces=true,                 
    numbers=left,                    
    numbersep=5pt,                  
    showspaces=false,                
    showstringspaces=false,
    showtabs=false,                  
    tabsize=2
}

\usepackage[most]{tcolorbox}
\usepackage{float}
\usepackage{xspace}
\tcbset{
  aibox/.style={
    width=474.18663pt,
    top=10pt,
    colback=white,
    colframe=black,
    colbacktitle=black,
    enhanced,
    center,
    attach boxed title to top left={yshift=-0.1in,xshift=0.15in},
    boxed title style={boxrule=0pt,colframe=white,},
  }
}
\newtcolorbox{AIbox}[2][]{aibox,title=#2,#1}

\def\huggingface{\raisebox{-1.5pt}{\includegraphics[height=1.05em]{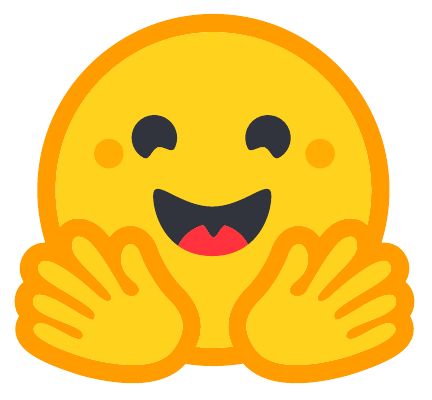}}}

\def\github{\raisebox{-1.5pt}{\includegraphics[height=1.05em]{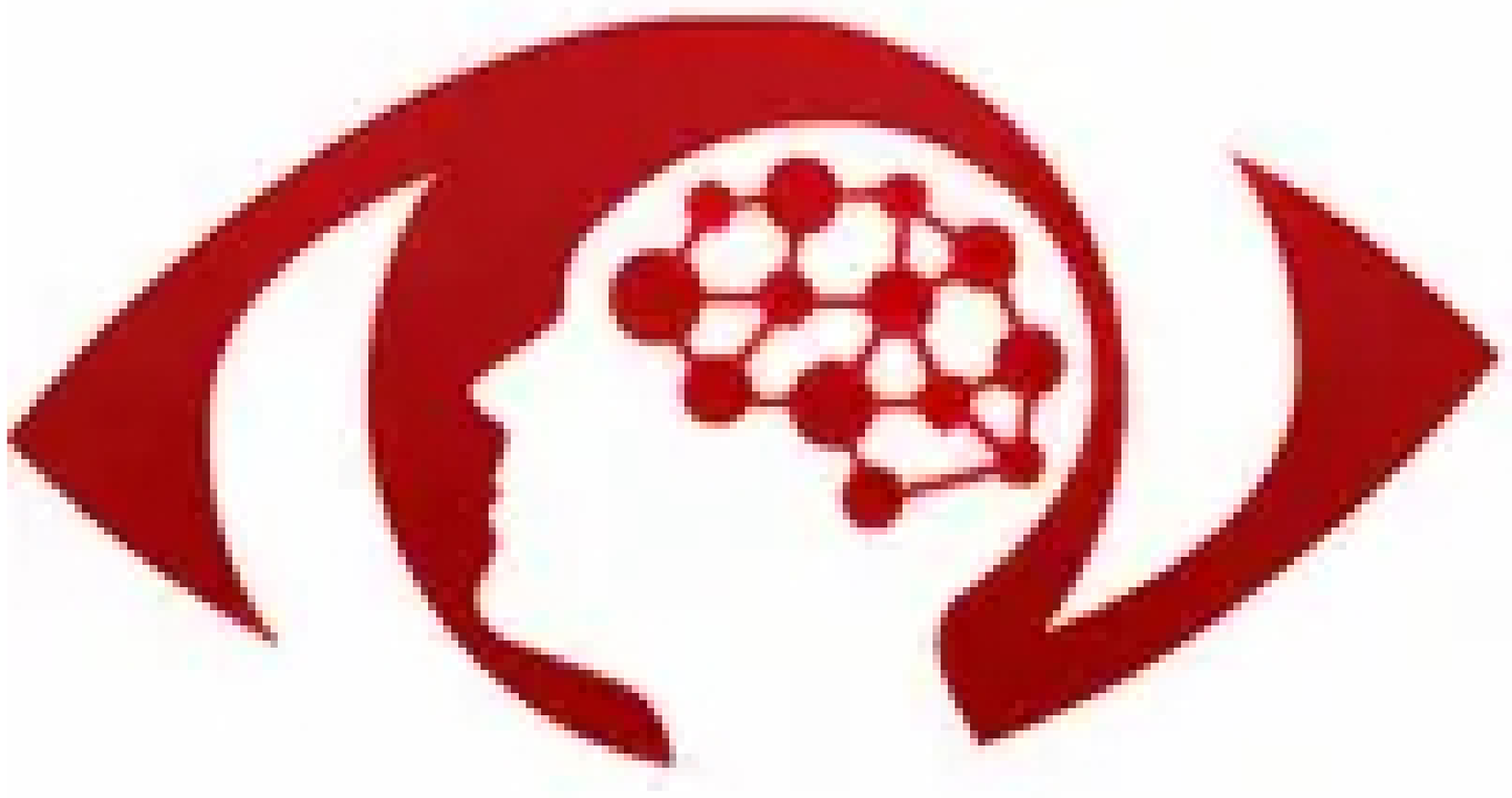}}}

\newcommand{\hflink}{https://huggingface.co/StanfordAIMI}

\newcommand{\ghlink}{https://stanford-aimi.github.io/srrg.html}

\title{Automated Structured Radiology Report Generation}

\author{
  Jean-Benoit Delbrouck$^{\spadesuit\heartsuit}$ \\
  \texttt{jbdel@stanford.edu} 
  \\\And
  Justin Xu$^{\spadesuit}$ 
  \\\And
  Johannes Moll$^{\spadesuit}$ 
   \\\AND
  Alois Thomas$^{\spadesuit}$ 
  \\\And
  Zhihong Chen$^{\spadesuit}$ 
  \\\And
    Sophie Ostmeier$^{\spadesuit}$ 
  \\\And
  Asfandyar Azhar$^{\spadesuit}$ 
   \\\AND
     Kelvin Zhenghao Li$^{\spadesuit}$ 
  \\\And
  Andrew Johnston$^{\spadesuit}$ 
  \\\And
    Christian Bluethgen$^{\spadesuit}$ 
   \\\AND
    Eduardo Reis$^{\spadesuit}$ 
  \\\And
  Mohamed Muneer$^{\spadesuit}$ 
  \\\And
    Maya Varma$^{\spadesuit}$ 
  \\\And
  Curtis Langlotz$^{\spadesuit}$ 
  \\\AND
  $^{\spadesuit}$ Stanford AIMI \hspace{0.2cm} $^{\heartsuit}$ HOPPR  \\
  \makebox[0pt][c]{%
    \begin{tabular}{rl}
        \huggingface & \url{\hflink}\\
        \github      & \url{\ghlink}
    \end{tabular}
    }
}

\begin{document}
\maketitle

\begin{abstract}
Automated radiology report generation from chest X-ray (CXR) images has the potential to improve clinical efficiency and reduce radiologists' workload. However, most datasets, including the publicly available MIMIC-CXR and CheXpert Plus, consist entirely of free-form reports, which are inherently variable and unstructured. This variability poses challenges for both generation and evaluation: existing models struggle to produce consistent, clinically meaningful reports, and standard evaluation metrics fail to capture the nuances of radiological interpretation. To address this, we introduce Structured Radiology Report Generation (SRRG), a new task that reformulates free-text radiology reports into a standardized format, ensuring clarity, consistency, and structured clinical reporting. We create a novel dataset by restructuring reports using large language models (LLMs) following strict structured reporting desiderata. Additionally, we introduce SRR-BERT, a fine-grained disease classification model trained on 55 labels, enabling more precise and clinically informed evaluation of structured reports. To assess report quality, we propose F1-SRR-BERT, a metric that leverages SRR-BERT’s hierarchical disease taxonomy to bridge the gap between free-text variability and structured clinical reporting. We validate our dataset through a reader study conducted by five board-certified radiologists and extensive benchmarking experiments.
\end{abstract}

\section{Introduction}
\label{sec:intro}
An important medical application of natural language generation (NLG) is the construction of assistive systems that take X-ray images of a patient and generate a textual report describing clinical observations in the images. This is a clinically important task, offering the potential to reduce the repetitive workload of radiologists and generally improve clinical communication~\cite{dunnick2008report,kahn2009toward}.\\

Since this task was first explored on chest X-ray (CXR) images, much of the related work, including exploring vanilla transformers~\cite{chen2020generating}, reinforcement learning algorithms~\cite{miura2021improving,delbrouck-etal-2022-improving}, and foundation models~\cite{chen2024chexagent, bannur2024maira}, has been conducted on two primary datasets: MIMIC-CXR~\cite{johnson2019mimic} and CheXpert Plus~\cite{chambon2024chexpert}. These datasets share notable similarities in terms of size, population diversity, and reporting style.\\

However, it is important to note that CXR reports themselves are typically free-form rather than structured by organ systems, primarily due to protocols, workflow efficiency, and the holistic nature of the necessary image interpretation \cite{weiss2008structured, bosmans2012structured}. This free-form style can pose unique challenges for automated report generation and clinical decision support as the variability in reporting styles often leads to inconsistencies in the way findings are described.\\

The need for more consistent, structured, or template-based radiology reporting is further reinforced by the difficulty faced by all proposed metrics in evaluating automated radiology report generation. Existing evaluation methods, ranging from standard NLG metrics such as BLEU~\cite{papineni2002bleu} and ROUGE~\cite{lin2004rouge} to clinical factuality-based metrics such as F1-RadGraph~\cite{delbrouck-etal-2022-improving}, RadFact~\cite{bannur2024maira}, or GREEN~\cite{ostmeier-etal-2024-green}, may struggle to capture the nuances of radiological interpretation due to the inherent diversity in reporting styles.\\

Given these limitations and observations, we introduce a new task, Structured Radiology Report Generation (SRRG, Section~\ref{sec:srrg}), aimed at transforming free-text radiology reports into a standardized format that enhances clarity and consistency through structured clinical documentation. To support this task, we present a new dataset derived from MIMIC-CXR and CheXpert Plus, where reports have been reformulated using large language models (LLMs) following strict desiderata for structured reporting. Additionally, we introduce SRR-BERT (Section~\ref{sec:disease_classif}), a novel disease classification model with 55 labels, designed to enable fine-grained automated evaluation. To further enhance the assessment of generated structured reports, we propose F1-SRR-BERT, a new metric that leverages SRR-BERT’s hierarchical disease taxonomy alongside a more precise evaluation paradigm made possible by the structured design of our task (Section~\ref{sec:evaluation metrics}). We validate our new datasets through a reader study (Section~\ref{sec:reader_study}) conducted by five board-certified radiologists, along with extensive experiments (Section~\ref{sec:benchmarking}).

\section{Structured Radiology Reporting} \label{sec:srrg}

\begin{figure*}[t]
    \centering
    \includegraphics[width=\textwidth]{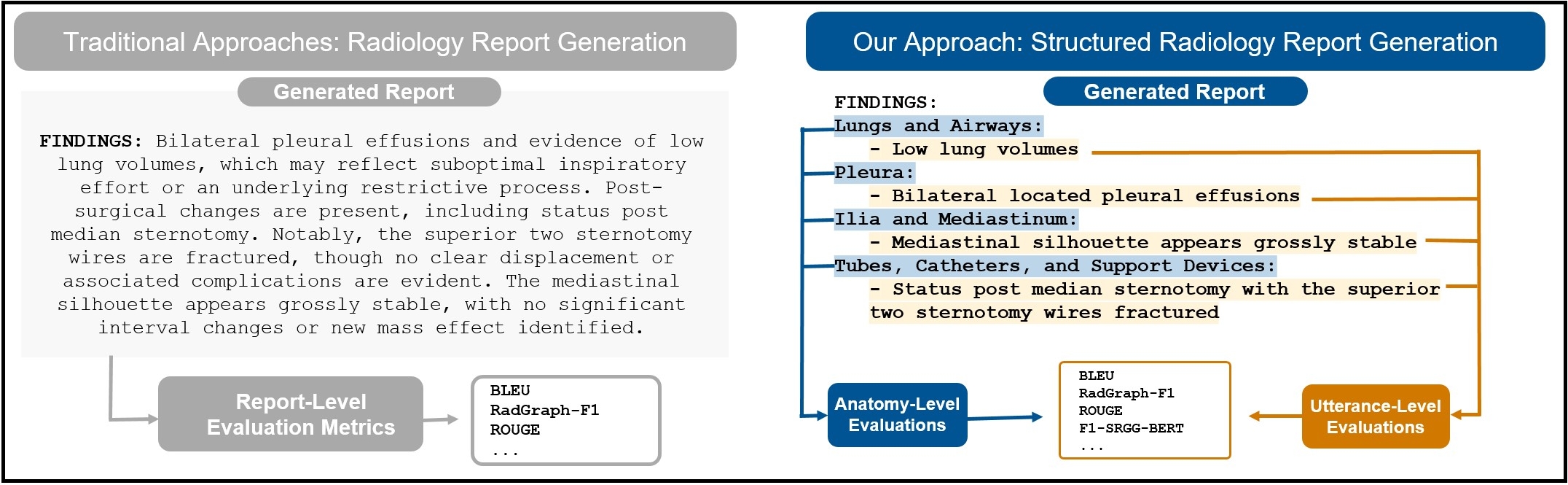}
    \caption{Comparison between traditional free-text radiology report generation (left) and our proposed Structured Radiology Report Generation (SRRG) approach (right). Traditional methods generate unstructured reports that vary in style and clarity, making automated evaluation challenging. In contrast, SRRG enforces a standardized format with anatomical section headers. This structured format enables more granular anatomy-level and utterance-level evaluations, including our proposed F1-SRR-BERT metric, which complements traditional report-level evaluation metrics.}
\end{figure*}

\subsection{Desiderata}
We define a structured radiology report as a report that follows a standardized format to ensure clarity and consistency. Such a report consists of distinct sections, each introduced by a section header followed by a colon, ensuring uniformity in presentation. The required sections include \textbf{Exam Type, History, Technique, Comparison, Findings, and Impression}. \\

The Findings section is organized under predefined anatomical headers, which are strictly limited to the following categories: \textbf{Lungs and Airways, Pleura, Cardiovascular, Hila and Mediastinum, Tubes, Catheters, and Support Devices, Musculoskeletal and Chest Wall, Abdominal, and Other}. Within each category, observations should be clearly listed using bullet points, and include all relevant positive and negative findings. \\

The Impression section summarizes the key findings in a numbered list, ranked from most to least clinically significant, ensuring that the most critical observations are highlighted effectively. \\

To maintain clarity and relevance, strict content guidelines need to be applied. References to previous studies or historical comparisons should be excluded, ensuring that the report reflects only the current examination. Identifiable information, including dates, surnames, first names, healthcare providers, vendors, and institutions, must be removed, although patient sex and age should be retained when provided. The content must strictly adhere to the defined structured sections, without extrapolating interpretations or introducing unrelated details. Additionally, only the specified anatomical headers may be used, ensuring a standardized report. The full prompt is available in Prompt~\ref{app:struct_prompt}.

\subsection{Dataset Creation} \label{sec:ssrg_data_creation}

Previous research has shown that GPT models can outperform traditional fine-tuned models in general summarization tasks by offering better factual consistency and reducing hallucinations~\cite{pu2023summarization}, achieve human-level performance in medical summarization of findings~\cite{van2024adapted}, and demonstrate strong capabilities in radiological error categorization~\cite{ostmeier-etal-2024-green}. Motivated by this, as well as by GPT-4’s "exceptional" performance across various medical benchmarks~\cite{nori2023capabilities}, we leverage LLMs to restructure the two largest publicly available chest X-ray datasets: MIMIC-CXR~\cite{johnson2019mimic} and CheXpert Plus~\cite{chambon2024chexpert}. The prompt used to rephrase the reports in accordance with our desiderata is provided in Prompt~\ref{app:disease tree}. This prompt was executed using GPT-4 "Turbo 1106 preview" via Azure services, with the account explicitly opted out of human review.

\subsection{Dataset Statistics}

We structured our dataset to align with the Radiology Report Generation (RRG) task by specifically mapping X-ray images to Findings (X-ray $\rightarrow$ Findings) and Impressions (X-ray $\rightarrow$ Impression). These setups correspond to our datasets, SRRG-Findings and SRRG-Impression, respectively. To construct the SRRG dataset, we combined MIMIC-CXR and CheXpert Plus and pooled them together to create our splits. Notably, SRRG-Impression is larger than SRRG-Findings, primarily because CheXpert predominantly contains Impression sections while often lacking Findings sections.

\begin{table}[h]
\centering
\renewcommand{\arraystretch}{1.2}   
\setlength{\tabcolsep}{4pt}         
\resizebox{\columnwidth}{!}{%
\begin{tabular}{lcc}
\toprule
\rowcolor{headerblue}
\textbf{Dataset} & \textbf{Split} & \textbf{Num. Examples} \\ 
\midrule
\multirow{4}{*}{SRRG-Impression} 
 & \cellcolor{gray!10}Train          & \cellcolor{gray!10}405,972 \\ 
 & Validate                         & 1,505 \\ 
 & \cellcolor{gray!10}Test           & \cellcolor{gray!10}2,219 \\ 
 & Test Reviewed                    & 231 \\ 
\cmidrule(lr){2-3}
 & \cellcolor{gray!10}\textbf{Total} & \cellcolor{gray!10}409,927 \\ 
\midrule
\multirow{4}{*}{SRRG-Findings} 
 & \cellcolor{gray!10}Train          & \cellcolor{gray!10}181,874 \\ 
 & Validate                         & 976 \\ 
 & \cellcolor{gray!10}Test           & \cellcolor{gray!10}1,459 \\ 
 & Test Reviewed                    & 233 \\ 
\cmidrule(lr){2-3}
 & \cellcolor{gray!10}\textbf{Total} & \cellcolor{gray!10}184,542 \\ 
\bottomrule
\end{tabular}%
}
\caption{Dataset statistics for SRRG-Impression and SRRG-Findings.}
\label{tab:srrg_datasets}
\end{table}

Lastly, we conducted a human review of 464 reports, sampled from the MIMIC-CXR test set and the CheXpert Plus validation set, with evaluations performed by five board-certified radiologists (Appendix~\ref{sec:reader_study}). Statistics of our datasets and splits are highlighted in Table~\ref{tab:srrg_datasets}.

\section{Disease Classification Models}\label{sec:disease_classif}
In this section, we introduce SRR-BERT, a novel model for fine-grained disease prediction that builds upon CheXbert to provide a more detailed assessment. Our approach extends the traditional set of 14 CheXbert disease labels to a set of 55 labels, covering a more granular hierarchy of pulmonary, pleural, cardiac, mediastinal, musculoskeletal, and abdominal findings, as well as more detailed support devices. This expanded taxonomy allows for more precise classification and evaluation of radiological abnormalities, enhancing the depth and accuracy of disease prediction.

\subsection{Desiderata} \label{sec:disease:desiderata}

To ensure \textbf{clarity, consistency, and clinical relevance}, our disease annotation framework follows the following key principles. Each finding must be mapped to \textbf{all relevant diseases} from a predefined list, allowing for zero, one, or multiple conditions. If no disease is present, the annotation explicitly states \textit{"No Finding"} to ensure systematic coverage.

Every disease is assigned a \textbf{status}—\textit{Present}, \textit{Absent}, or \textit{Uncertain}—capturing clinical uncertainty and preventing over-assumptions. For example: 

\begin{quote}
\textit{Right perihilar consolidation, likely atypical edema, with pneumonia as a differential diagnosis.}
\end{quote}

is annotated as:

\begin{quote}
\texttt{=> Perihilar airspace opacity (Present)}\\
\texttt{=> Edema (Uncertain)}\\
\texttt{=> Pneumonia (Uncertain)}
\end{quote}

The selected diseases and their hierarchical structure are detailed in Prompt~\ref{app:disease tree}. This disease tree has been validated by a board-certified radiologist. While the first level of the hierarchical structure corresponds to the Anatomical Headers / Category, the lowest level is referred to as tree "leaves", and "upper" labels denote the item one-level above "leaves". Appendix~\ref{app:disease_breakdown} shows a dataset breakdown of each of the "upper" labels.

\subsection{Dataset Creation}

We annotate all utterances in our SRRG dataset, where an utterance is defined as either a single-sentence finding or a numbered impression. This process results in 1,562,277 unique utterances. To ensure consistency in annotation, we follow the guidelines outlined in Section~\ref{sec:disease:desiderata} and craft the structured annotation template accordingly provided in Prompt~\ref{app:disease_prompt}. \\

To validate the correctness of the assigned labels, we employ both automated and human reviews. The automated review follows a mixture-of-experts approach, where each utterance is processed using three different GPT models: GPT-4 Turbo (2024-04-09), GPT-4 Turbo 1106 Preview, and GPT-4o (2024-08-06). The final labels for each utterance are determined by selecting the diseases that appear in at least two out of the three model outputs. This ensures robustness and reduces inconsistencies in the predictions. If an utterance has no labels, we discard it. We ultimately obtain a total of 1,506,158 valid utterances (as detailed in Section \ref{sec:utterance_dataset_stats})

\subsection{Dataset Statistics}\label{sec:utterance_dataset_stats}
The dataset comprises 1,506,158 utterances annotated with 1,782,983 labels, averaging 1.18 labels per utterance. Among all utterances, 905,764 correspond to positive findings (i.e., not labeled as "No Finding"), with these having an average of 1.31 labels per utterance. \\

\begin{table}[h]
\centering
\renewcommand{\arraystretch}{1.2}   
\setlength{\tabcolsep}{4pt}         
\resizebox{\columnwidth}{!}{%
\begin{tabular}{lcc}
\toprule
\rowcolor{headerblue}
\textbf{Dataset} & \textbf{Split} & \textbf{Num. Examples} \\
\midrule
\multirow{4}{*}{StructUtterances}
 & \cellcolor{gray!10}Train          & \cellcolor{gray!10}1,203,332 \\
 & Validate                         & 150,417 \\
 & \cellcolor{gray!10}Test           & \cellcolor{gray!10}150,417 \\
 & Test Reviewed                    & 1,609 \\
\cmidrule(lr){2-3}
 & \cellcolor{gray!10}\textbf{Total} & \cellcolor{gray!10}1,506,158 \\
\bottomrule
\end{tabular}%
}
\caption{Dataset statistics for StructUtterances.}
\label{tab:structutterance_dataset}
\end{table}

The test-reviewed split was evaluated by five board-certified radiologists (Appendix~\ref{sec:reader_study}) and includes utterances extracted from the reports in the test-reviewed split of our SRRG dataset (Table~\ref{tab:srrg_datasets}).

\section{Benchmarking}~\label{sec:benchmarking}

\subsection{Disease Classification Models}
To benchmark disease classification, we fine-tune CXR-BERT~\cite{Boecking_2022} on weakly-labeled utterances in the StructUtterances dataset under four experimental settings. First, we set aside the status annotations (i.e., Present, Absent, Uncertain) and only classify the "leaves" and "upper" labels. We then integrate the three statuses by creating a separate class for each combination, yielding "leaves with statuses" and "upper with statuses".

The benchmarking results for the disease classification models demonstrate strong overall performance on the reviewed test split, with F1 scores exceeding 0.75 for most classes. However, as is typical in classification tasks, rare labels posed a challenge. For the model operating at the "leaves" level, the overall F1 score was 0.836, with the three best-performing labels being "No Finding" (F1 = 0.83, n=452), "Simple Pleural Effusion" (F1 = 0.93, n=174), and "Atelectasis" (F1 = 0.94, n=131). Noticeably poor-performing classes include "Air space opacity-multifocal" (F1 = 0.62, n=60) and "Suboptimal central line" (F1 = 0.19, n=29). At the "upper" level with reduced granularity, our model achieved an overall F1 score of 0.839, with top-three performing labels being "No Finding" (F1 = 0.82, n=452), "Consolidation" (F1 = 0.89, n=215), and "Pleural Effusion" (F1 = 0.94, n=185).\\

When incorporating status annotations, performance declined slightly due to the number of labels effectively being tripled. The "leaves with statuses" model yielded an F1 score of 0.794, while the "upper with statuses" model achieved an F1 score of 0.795. In both cases, "No Finding" remained a strong performer (F1 = 0.82), while disease-specific labels such as "Simple Pleural Effusion (Present)" (F1 = 0.91, n=96) and "Cardiomegaly (Present)" (F1 = 0.98, n=82) performed very well. However, some uncertain findings, such as "Consolidation (Uncertain)" (F1 = 0.82, n=95), demonstrated slightly lower scores, reflecting the intrinsic difficulty of differentiating between ambiguous disease states.

\begin{table}[h]
\centering
\small
\renewcommand{\arraystretch}{1.2}   
\setlength{\tabcolsep}{4pt}         
\rowcolors{3}{gray!10}{white}       
\resizebox{\columnwidth}{!}{%
\begin{tabular}{lcccc}
\toprule
\rowcolor{headerblue}
 & \textbf{Precision} & \textbf{Recall} & \textbf{F1-Score} & \textbf{Support} \\
\midrule
& \multicolumn{4}{c}{\textbf{\textit{Leaves}}} \\
\cmidrule(lr){2-5}
Micro Avg    & \textbf{0.85} & 0.82 & \textbf{0.84} & 1,644 \\
Macro Avg    & 0.63          & 0.53 & 0.55          & 1,644 \\
Weighted Avg & \textbf{0.85} & 0.82 & 0.82          & 1,644 \\
Samples Avg  & 0.84          & \textbf{0.84} & \textbf{0.84} & 1,644 \\

\midrule
& \multicolumn{4}{c}{\textbf{\textit{Upper}}} \\
\cmidrule(lr){2-5}
Micro Avg    & 0.85          & 0.83 & \textbf{0.84} & 1,588 \\
Macro Avg    & 0.70          & 0.62 & 0.65          & 1,588 \\
Weighted Avg & \textbf{0.87} & 0.83 & 0.83          & 1,588 \\
Samples Avg  & 0.85          & \textbf{0.84} & \textbf{0.84} & 1,588 \\

\midrule
& \multicolumn{4}{c}{\textbf{\textit{Leaves with Statuses}}} \\
\cmidrule(lr){2-5}
Micro Avg    & \textbf{0.81} & 0.78 & \textbf{0.80} & 1,644 \\
Macro Avg    & 0.31          & 0.27 & 0.28          & 1,644 \\
Weighted Avg & 0.79          & 0.78 & 0.77          & 1,644 \\
Samples Avg  & 0.80          & \textbf{0.80} & 0.79          & 1,644 \\

\midrule
& \multicolumn{4}{c}{\textbf{\textit{Upper with Statuses}}} \\
\cmidrule(lr){2-5}
Micro Avg    & \textbf{0.81} & 0.79 & \textbf{0.80} & 1,574 \\
Macro Avg    & 0.41          & 0.38 & 0.38          & 1,574 \\
Weighted Avg & 0.79          & 0.79 & 0.78          & 1,574 \\
Samples Avg  & 0.80          & \textbf{0.80} & \textbf{0.80} & 1,574 \\

\bottomrule
\end{tabular}%
}
\caption{Benchmark results for disease classification on the test\_reviewed split. Highest scores are in bold.}
\label{tab:srrg_benchmark}
\end{table}

\subsubsection{Comparison to CheXbert}
We compare our models to CheXbert as they both aim to accomplish the same task of disease classification. Given the more restricting label set of CheXbert, we first filter the reviewed test set to only include utterances with a label that is mappable to CheXbert classes. This mapping between label spaces was conducted after consulting a combination of web sources, a clinician, and GPT-4o. However, some degree of overlap and ambiguity remains (Section~\ref{sec:limitations}).\\

Using structured utterances as input, we first derive CheXbert labels using the author-provided CheXbert model checkpoint. Using SRR-BERT, we then compute labels at both the "leaves" level and the "upper" level and map them to the 14 classes used by CheXbert. Table~\ref{tab:chexbert_srrg} illustrates the direct comparison of model performances, where SRR-BERT outperformed CheXbert in both settings (0.80 vs. 0.61 when "leaves" were used for the mapping, and 0.83 vs. 0.47 when "upper" labels were used for the mapping).\\

We acknowledge that SRR-BERT was trained on structured utterances while CheXbert was not, which may skew the comparison. Hence, we also leverage the unstructured full-length reports as input. SRR-BERT outperforms CheXbert when using "upper" labels to map to CheXbert classes, and exhibits only slightly lower F1 when using "leaves". This demonstrates the robustness of SRR-BERT models as they can accommodate texts of varying lengths and complexity, from short utterances to full-length reports.







\begin{table}[ht]
\centering
\small
\renewcommand{\arraystretch}{1.2}   
\setlength{\tabcolsep}{4pt}         
\rowcolors{3}{gray!10}{white}       
\resizebox{\columnwidth}{!}{%
\begin{tabular}{lcccc}
\toprule
\rowcolor{headerblue}
 & \textbf{Precision} & \textbf{Recall} & \textbf{F1-Score} & \textbf{Support} \\
\midrule
\multicolumn{5}{c}{\textbf{\textit{Mapped with Leaves}}} \\
\midrule
\textbf{Utterances} & & & & \\
CheXbert  & 0.69 & 0.64 & 0.65 & 1,759 \\
SRR-BERT  & \textbf{0.88} & \textbf{0.82} & \textbf{0.84} & 1,759 \\
\midrule
\textbf{Full Reports} & & & & \\
CheXbert  & 0.73 & \textbf{0.59} & \textbf{0.62} & 260 \\
SRR-BERT  & \textbf{0.84} & 0.48 & 0.58 & 260 \\
\midrule
\multicolumn{5}{c}{\textbf{\textit{Mapped with Upper}}} \\
\midrule
\textbf{Utterances} & & & & \\
CheXbert  & 0.70 & 0.48 & 0.50 & 2,004 \\
SRR-BERT  & \textbf{0.90} & \textbf{0.84} & \textbf{0.86} & 2,004 \\
\midrule
\textbf{Full Report} & & & & \\
CheXbert  & 0.80 & 0.49 & 0.56 & 278 \\
SRR-BERT  & \textbf{0.89} & \textbf{0.60} & \textbf{0.70} & 278 \\
\bottomrule
\end{tabular}%
}
\caption{Weighted average performance comparison for CheXbert and SRR-BERT using “leaves” and “upper” mappings to 14 CheXbert classes on the test\_reviewed split. Highest scores are in bold.}
\label{tab:chexbert_srrg}
\end{table}

\subsection{Structured RRG}

\subsubsection{Evaluation Metrics} \label{sec:evaluation metrics}
To ensure consistency with prior work in "traditional" RRG, we report BLEU~\cite{papineni2002bleu}, BERTScore~\cite{zhang2019bertscore}, ROUGE-L~\cite{lin2004rouge}, and F1-RadGraph~\cite{delbrouck-etal-2022-improving}. Additionally, we introduce F1-SRR-BERT, a new metric leveraging our SRR-BERT model (Section~\ref{sec:disease_classif}), which is trained to predict abnormalities across 55 diseases based on CXR utterances. \\

F1-SRR-BERT measures the F1-Score between SRR-BERT's predictions on the generated structured report and the corresponding reference structured reports. This score has two variants: (1) \textit{leaves prediction}, which classifies diseases at the finest granularity (55 labels from the disease tree in Prompt~\ref{app:disease tree}), and (2) \textit{upper-level prediction}, which groups diseases into 25 broader categories for a coarser classification. These broader categories are the level right above the "leaves".\\

An additional consideration in our evaluation is that utterances can be assessed in either an \textit{aligned} or \textit{unaligned} setting across all previously mentioned metrics. In the \textit{aligned} setting, utterances are evaluated in the order they appear under an organ system header or by their numerical order in the impression section (i.e., generated impression one is compared to reference impression one). In contrast, the \textit{unaligned} setting evaluates utterances as a set—comparing all findings under an organ system or all numbered impressions as a block against the reference. This unaligned approach allows us to assess whether the model prioritizes findings and impressions from the most to the least clinically relevant. Finally, we assign a score of 0 for missing references sections and extra predicted sections in findings.

\subsubsection{Results}
We benchmark four distinct models: MAIRA-2~\cite{bannur2024maira}, CheXagent~\cite{chen2024chexagent}, CheXpert-Plus~\cite{chambon2024chexpert}, and RaDialog~\cite{pellegrini2023radialog}. These models vary in size, architecture, and reported performance.
\begin{table*}[ht]
\centering
\renewcommand{\arraystretch}{1.2}     
\setlength{\tabcolsep}{6pt}           
\rowcolors{3}{gray!10}{white}         
\resizebox{\textwidth}{!}{%
\begin{tabular}{llccccccc}
\toprule
\rowcolor{headerblue}
\multicolumn{2}{c}{\textbf{SRRG-Impression (unaligned)}} &
  \multicolumn{4}{c}{\textbf{Traditional Metrics}} &
  \multicolumn{3}{c}{\textbf{F1-SRR-BERT}} \\
\cmidrule(lr){1-2}\cmidrule(lr){3-6}\cmidrule(lr){7-9}
\rowcolor{headerblue}
\textbf{Model} & \textbf{Split} &
  \textbf{BLEU} & \textbf{ROUGE-L} & \textbf{BERTScore} & \textbf{F1-RadGraph} &
  \textbf{Precision} & \textbf{Recall} & \textbf{F1-Score} \\
\midrule
CheXagent       & Validate      & 7.86  & 28.94 & 60.55 & 20.62 & 50.02 & 56.32 & 50.60 \\
CheXagent       & Test          & \textbf{6.95}  & \textbf{27.18} & \textbf{61.51} & \textbf{19.70} & \textbf{49.78} & 56.47 & \textbf{50.63} \\
CheXagent       & Test Reviewed & 4.68  & 26.10 & 59.70 & 18.33 & 45.24 & \textbf{56.70} & 48.64 \\
\midrule
CheXpert-Plus   & Validate      & 16.86 & 33.42 & 62.74 & 27.74 & 54.40 & 51.26 & 50.26 \\
CheXpert-Plus   & Test          & \textbf{14.84} & \textbf{28.01} & \textbf{60.76} & \textbf{22.14} & \textbf{48.74} & 47.60 & \textbf{46.48} \\
CheXpert-Plus   & Test Reviewed & 14.07 & 26.79 & 59.21 & 18.89 & 43.46 & \textbf{48.15} & 44.56 \\
\midrule
MAIRA-2         & Validate      & 9.66  & 31.50 & 62.84 & 23.21 & 52.53 & 61.16 & 54.46 \\
MAIRA-2         & Test          & \textbf{8.12}  & \textbf{27.82} & \textbf{62.30} & \textbf{20.37} & \textbf{48.72} & \textbf{57.91} & \textbf{50.36} \\
MAIRA-2         & Test Reviewed & 5.28  & 26.61 & 60.79 & 19.08 & 44.80 & 57.69 & 47.97 \\
\midrule
RaDialog        & Validate      & 5.35  & 23.93 & 57.74 & 15.27 & 39.80 & 52.41 & 40.70 \\
RaDialog        & Test          & 3.32  & \textbf{21.59} & \textbf{57.48} & \textbf{12.32} & \textbf{37.30} & 50.59 & \textbf{39.22} \\
RaDialog        & Test Reviewed & \textbf{3.33}  & 19.95 & 54.82 & 10.26 & 33.65 & \textbf{50.71} & 36.39 \\
\bottomrule
\end{tabular}%
}
\caption{Model scores on different splits of our \textbf{SRRG-impression} dataset. Traditional metrics (BLEU, ROUGE-L, BERTScore, F1-RadGraph) are shown as percentages. F1-SRR-BERT scores (weighted averages for utterance-level diseases Precision, Recall, and F1-Score). Bold indicates the best score per model group on the Test vs.\ Test Reviewed splits.}
\label{tab:model_scores_main_impression}
\end{table*}

\textbf{Impression.} \quad Table~\ref{tab:model_scores_main_impression} shows the performance of various models in generating impressions (evaluated without alignment), revealing that models tend to score higher in this task than in free-form impression generation. Notably, CheXpert-Plus stands out as the best performer on the SRRG-Impression dataset. On the test split, it achieves the highest traditional metric scores, with a BLEU of 14.84, ROUGE-L of 28.01, and F1-RadGraph of 22.14, while also registering the highest utterance-level precision at 58.99. Although CheXagent and MAIRA-2 excel in BERTScore and Recall respectively, CheXpert-Plus consistently delivers superior performance across both traditional and SRRG metrics.

\begin{table}[ht]
\centering
\renewcommand{\arraystretch}{1.2}   
\setlength{\tabcolsep}{6pt}         
\rowcolors{3}{gray!10}{white}       
\resizebox{\columnwidth}{!}{%
\begin{tabular}{lcccc}
\toprule
\rowcolor{headerblue}
\textbf{Split} & \textbf{BLEU} & \textbf{ROUGE-L} & \textbf{BERTScore} & \textbf{F1-RadGraph} \\
\midrule
\multicolumn{5}{c}{\textbf{SRRG-Impression}} \\
Validate       & 7.61 \textcolor{red}{$\downarrow$9.25}   & 23.35 \textcolor{red}{$\downarrow$10.07} & 39.95 \textcolor{red}{$\downarrow$22.79} & 16.68 \textcolor{red}{$\downarrow$11.06}   \\
Test           & 3.78 \textcolor{red}{$\downarrow$11.06}  & 16.77 \textcolor{red}{$\downarrow$11.24} & 36.35 \textcolor{red}{$\downarrow$24.41} & 10.23 \textcolor{red}{$\downarrow$11.91}  \\
Test Reviewed  & 3.63 \textcolor{red}{\textbf{$\downarrow$10.44}} & 16.89 \textcolor{red}{\textbf{$\downarrow$9.90}}  & 38.82 \textcolor{red}{\textbf{$\downarrow$20.39}} & 10.42 \textcolor{red}{\textbf{$\downarrow$8.47}}   \\
\midrule
\multicolumn{5}{c}{\textbf{SRRG-Findings}} \\
Validate       & 3.77 \textcolor{red}{$\downarrow$0.35}   & 19.23 \textcolor{red}{$\downarrow$1.67}   & 26.81 \textcolor{red}{$\downarrow$4.77}   & 14.23 \textcolor{red}{$\downarrow$2.72}    \\
Test           & 3.21 \textcolor{red}{\textbf{$\downarrow$0.30}}  & 16.89 \textcolor{red}{\textbf{$\downarrow$2.08}}  & 25.83 \textcolor{red}{\textbf{$\downarrow$5.67}}  & 12.31 \textcolor{red}{\textbf{$\downarrow$2.68}}   \\
Test Reviewed  & 3.45 \textcolor{red}{$\downarrow$0.51}  & 16.27 \textcolor{red}{$\downarrow$2.45}  & 24.93 \textcolor{red}{$\downarrow$6.40}  & 11.68 \textcolor{red}{$\downarrow$3.21}    \\
\bottomrule
\end{tabular}%
}
\caption{Updated scores for the CheXpert-Plus model using the “\textbf{aligned}” settings. The differences from the unaligned settings (Tables~\ref{tab:model_scores_main_impression} and~\ref{tab:model_scores_main_findings}) are shown in red. For each section, the smaller drop between the Test and Test Reviewed splits is highlighted in bold.}
\label{tab:aligned}
\end{table}

\textbf{Findings.} \quad In the SRRG-Findings (unaligned) setting (Table~\ref{tab:model_scores_main_findings}), traditional metric scores are generally lower than in the SRRG-Impression setting, indicating that generating structured findings is more challenging than producing impressions. For findings, CheXpert-Plus achieves moderate scores on validation (e.g., BLEU 4.12, ROUGE-L 20.90, BERTScore 31.58, F1-RadGraph 16.95), while CheXagent and MAIRA-2 show similar patterns with slight drops from validation to test splits. Category scores---reflecting the correct prediction of organ section headers---are consistently high (around 75–78\%) across models. In contrast, the impression results reveal substantially higher traditional metrics, with CheXagent and CheXpert-Plus achieving BLEU scores above 14 and BERTScores in the low 60s, suggesting that the impression task yields more polished, concise outputs. Overall, these results highlight that while all models struggle with the detailed nature of findings, they perform significantly better when generating shorter, impression-style summaries. \\

\begin{table*}[t]
\centering
\renewcommand{\arraystretch}{1.2}     
\setlength{\tabcolsep}{6pt}           
\rowcolors{3}{gray!10}{white}         
\resizebox{\textwidth}{!}{%
\begin{tabular}{llccccccccccc}
\toprule
\rowcolor{headerblue}
\multicolumn{2}{c}{\textbf{SRRG-Findings (unaligned)}} &
  \multicolumn{4}{c}{\textbf{Traditional Metrics}} &
  \multicolumn{6}{c}{\textbf{F1-SRR}} \\
\cmidrule(lr){1-2} \cmidrule(lr){3-6} \cmidrule(lr){7-12}
\rowcolor{headerblue}
\multicolumn{2}{c}{} &
  \multicolumn{4}{c}{} &
  \multicolumn{3}{c}{\textbf{F1-SRR-BERT}} &
  \multicolumn{3}{c}{\textbf{Category}} \\
\cmidrule(lr){7-9} \cmidrule(lr){10-12}
\rowcolor{white}
\textbf{Model} & \textbf{Split} &
  \textbf{BLEU} & \textbf{ROUGE-L} & \textbf{BERTScore} & \textbf{F1-RadGraph} &
  \textbf{Precision} & \textbf{Recall} & \textbf{F1-Score} &
  \textbf{Precision} & \textbf{Recall} & \textbf{F1-Score} \\
\midrule
CheXagent       & Validate       & 1.93 & 19.72 & 29.58 & 15.35 & 42.86 & 44.04 & 41.88 & 75.98 & 77.16 & 74.70 \\
CheXagent       & Test           & 1.80 & 19.65 & 31.65 & 15.41 & 43.22 & 42.07 & 41.13 & \textbf{77.12} & 82.56 & \textbf{77.90} \\
CheXagent       & Test Reviewed  & \textbf{2.38} & \textbf{19.88} & \textbf{32.48} & \textbf{16.04} & \textbf{44.56} & \textbf{42.53} & \textbf{41.73} & 75.26 & \textbf{85.22} & 77.40 \\
\midrule
CheXpert-Plus   & Validate       & 4.12 & 20.90 & 31.58 & 16.95 & 44.28 & 43.19 & 42.08 & 72.10 & 85.45 & 76.52 \\
CheXpert-Plus   & Test           & 3.51 & \textbf{18.97} & \textbf{31.50} & \textbf{14.99} & \textbf{42.79} & \textbf{40.08} & \textbf{39.85} & \textbf{72.84} & 86.17 & 77.18 \\
CheXpert-Plus   & Test Reviewed  & \textbf{3.96} & 18.72 & 31.33 & 14.89 & 42.78 & 39.10 & 39.28 & 71.63 & \textbf{88.71} & \textbf{77.24} \\
\midrule
MAIRA-2         & Validate       & 6.32 & 29.00 & 39.38 & 25.62 & 49.66 & 49.66 & 49.66 & 78.21 & 86.24 & 80.52 \\
MAIRA-2         & Test           & \textbf{3.39} & \textbf{23.15} & \textbf{35.44} & \textbf{19.03} & \textbf{43.65} & \textbf{43.65} & \textbf{43.65} & \textbf{75.64} & 86.23 & \textbf{79.03} \\
MAIRA-2         & Test Reviewed  & 2.26 & 20.55 & 32.87 & 16.90 & 42.36 & 42.36 & 42.36 & 72.25 & \textbf{88.90} & 77.79 \\
\midrule
RaDialog        & Validate       & 1.47 & 18.23 & 28.67 & 13.92 & 40.15 & 39.63 & 39.08 & 70.12 & 70.48 & 69.33 \\
RaDialog        & Test           & 1.28 & 17.53 & \textbf{29.07} & \textbf{13.82} & 38.42 & 38.10 & 37.89 & 69.48 & 70.12 & \textbf{69.76} \\
RaDialog        & Test Reviewed  & \textbf{1.42} & \textbf{17.60} & 28.90 & 13.75 & \textbf{38.95} & \textbf{38.30} & \textbf{38.05} & \textbf{69.90} & \textbf{70.22} & 69.85 \\
\bottomrule
\end{tabular}%
}
\caption{Model scores on different splits of our \textbf{SRRG-Findings} dataset. Traditional Metrics include BLEU, ROUGE-L, BERTScore, and F1-RadGraph. F1-SRR-BERT metrics (weighted averages) are evaluated for Diseases and for Category (organ section headers). Bold indicates the best score per model group on the Test vs.\ Test Reviewed splits.}
\label{tab:model_scores_main_findings}
\end{table*}

\textbf{Alignment.} \quad As expected, generating impressions and findings that align with the ground-truth is challenging, as demonstrated by CheXpert-Plus' scores (Table~\ref{tab:aligned}). This challenge is even more pronounced in the impression setting, which typically contains more utterances than organ sections.

\begin{table}[ht]
\centering
\renewcommand{\arraystretch}{1.2}   
\setlength{\tabcolsep}{4pt}         
\rowcolors{2}{gray!10}{white}       
\resizebox{\columnwidth}{!}{%
\begin{tabular}{lccc}
\toprule
\rowcolor{headerblue}
\textbf{Organ} & \textbf{Precision} & \textbf{Recall} & \textbf{F1-Score} \\
\midrule
Pleura                                & 54.53 & 40.28 & 44.23 \\
Abdominal                             & 10.53 & 10.53 & 10.53 \\
Hila and Mediastinum                  & 22.26 & 21.58 & 21.69 \\
Other                                 &  3.69 &  3.42 &  3.39 \\
Lungs and Airways                     & 41.85 & 40.41 & 38.32 \\
Cardiovascular                        & 63.78 & 58.73 & 59.78 \\
Musculoskeletal and Chest Wall        & 45.99 & 43.91 & 44.29 \\
Tubes, Catheters, and Support Devices & 51.27 & 54.94 & 50.56 \\
\bottomrule
\end{tabular}%
}
\caption{Organ-level F1-SRRG-BERT weighted-average scores for CheXpert-Plus on the test-reviewed split.}
\label{tab:chexpert_plus_organs}
\end{table}

Table~\ref{tab:chexpert_plus_organs} reveals marked heterogeneity in CheXpert-Plus's organ-specific performance. The model is most reliable for cardiovascular structures, with an F1 score of roughly 60, and for hardware-related findings (``Tubes, Catheters, and Support Devices''), where the score is about 51; pleural and musculoskeletal regions follow, each in the mid-40s. Performance drops substantially for lung parenchyma and airways, which score around 38, and is weakest for abdominal findings (about 11) and the miscellaneous ``Other'' category (around 3). These disparities suggest that CheXpert-Plus excels when imaging cues are distinct or well-represented in the training data, but struggles with rarer or more heterogeneous organ systems.

\textbf{OOD.} \quad Finally, we perform an out-of-distribution (OOD) evaluation using the HOPPR test set, which consists of 1,300 samples sourced from the HOPPR Platform. These samples come from data providers across eight U.S. states. Each report in the set contains at least one confirmed positive finding, including conditions such as Acute Rib Fracture, Air Space Opacity, Cardiomegaly, Lung Nodule or Mass, Non-Acute Rib Fracture, Pleural Fluid, Pneumothorax, or Pulmonary Artery Enlargement. When tested on this new, out-of-distribution dataset, all three public models exhibit a typical domain shift: their lexical metrics—such as BLEU, ROUGE-L, and BERTScore (drop by 3 to 15 points). However, structure-aware metrics remain much more stable. RadGraph F1 decreases by only about 1.5 points, and interestingly, disease-level F1 using SRR-BERT for the Findings section actually increases by 0.8 to 2.4 points. Performance on organ-category labels also improves, rising by 3 to 4 points. The main weakness lies in generating the Impression section, where models lose between 13 and 18 points.

\section{Conclusion}

We presented Structured Radiology Report Generation (SRRG), a new task reformulating free-text CXR reports into standardized templates to improve clarity and enable more precise evaluation. To support SRRG, we introduce a large-scale dataset with clinically validated structured reports and SRR-BERT, a 55-label disease classifier trained on fine-grained radiological findings. We further propose F1-SRR-BERT, a metric leveraging SRR-BERT’s hierarchical labels to capture clinically meaningful variations. Our reader study, conducted by board-certified radiologists, confirms the quality of both the structured reports and annotated disease labels. Benchmark experiments show that SRRG improves consistency compared to existing free-form generation methods.

\begin{table*}[t]
\centering
\renewcommand{\arraystretch}{1.2}     
\setlength{\tabcolsep}{6pt}           
\rowcolors{3}{gray!10}{white}         
\resizebox{\textwidth}{!}{%
\begin{tabular}{llccccccccccc}
\toprule
\rowcolor{headerblue}
\multicolumn{2}{c}{\textbf{SRRG-Findings (unaligned)}} &
  \multicolumn{4}{c}{\textbf{Traditional Metrics}} &
  \multicolumn{6}{c}{\textbf{F1-SRR}} \\
\cmidrule(lr){1-2}\cmidrule(lr){3-6}\cmidrule(lr){7-12}
\rowcolor{headerblue}
\multicolumn{2}{c}{} &
  \multicolumn{4}{c}{} &
  \multicolumn{3}{c}{\textbf{F1-SRR-BERT}} &
  \multicolumn{3}{c}{\textbf{Category}} \\
\cmidrule(lr){7-9}\cmidrule(lr){10-12}
\rowcolor{white} 
\textbf{Model} & \textbf{Split} &
  \textbf{BLEU} & \textbf{ROUGE-L} & \textbf{BERTScore} & \textbf{F1-RadGraph} &
  \textbf{Precision} & \textbf{Recall} & \textbf{F1-Score} &
  \textbf{Precision} & \textbf{Recall} & \textbf{F1-Score} \\
\midrule
CheXagent & Test (OOD) & 
  3.90\,\textcolor{teal}{$\uparrow$2.10} &
  16.50\,\textcolor{red}{$\downarrow$3.15} &
  28.10\,\textcolor{red}{$\downarrow$3.55} &
  13.70\,\textcolor{red}{$\downarrow$1.71} &
  42.70\,\textcolor{red}{$\downarrow$0.52} &
  44.10\,\textcolor{teal}{$\uparrow$2.03} &
  43.38\,\textcolor{teal}{$\uparrow$2.25} &
  77.70\,\textcolor{teal}{$\uparrow$0.58} &
  87.50\,\textcolor{teal}{$\uparrow$4.94} &
  82.30\,\textcolor{teal}{$\uparrow$4.40} \\
CheXpert-Plus & Test (OOD) &
  6.10\,\textcolor{teal}{$\uparrow$2.59} &
  15.84\,\textcolor{red}{$\downarrow$3.13} &
  28.00\,\textcolor{red}{$\downarrow$3.50} &
  13.31\,\textcolor{red}{$\downarrow$1.68} &
  42.28\,\textcolor{red}{$\downarrow$0.51} &
  42.28\,\textcolor{teal}{$\uparrow$2.20} &
  42.28\,\textcolor{teal}{$\uparrow$2.43} &
  73.50\,\textcolor{teal}{$\uparrow$0.66} &
  91.56\,\textcolor{teal}{$\uparrow$5.39} &
  80.91\,\textcolor{teal}{$\uparrow$3.73} \\
MAIRA-2 & Test (OOD) &
  5.90\,\textcolor{teal}{$\uparrow$2.51} &
  20.00\,\textcolor{red}{$\downarrow$3.15} &
  31.90\,\textcolor{red}{$\downarrow$3.54} &
  17.30\,\textcolor{red}{$\downarrow$1.73} &
  43.10\,\textcolor{red}{$\downarrow$0.55} &
  45.90\,\textcolor{teal}{$\uparrow$2.25} &
  44.45\,\textcolor{teal}{$\uparrow$0.80} & 
  76.20\,\textcolor{teal}{$\uparrow$0.56} &
  91.80\,\textcolor{teal}{$\uparrow$5.57} &
  82.80\,\textcolor{teal}{$\uparrow$3.77} \\
\midrule
\rowcolor{headerblue}
\multicolumn{2}{c}{\textbf{SRRG-Impression (unaligned)}} &
  \multicolumn{4}{c}{\textbf{Traditional Metrics}} &
  \multicolumn{3}{c}{\textbf{F1-SRR-BERT}} &
  \multicolumn{3}{c}{} \\ 
\cmidrule(lr){1-2}\cmidrule(lr){3-6}\cmidrule(lr){7-9}
\rowcolor{white} 
\textbf{Model} & \textbf{Split} &
  \textbf{BLEU} & \textbf{ROUGE-L} & \textbf{BERTScore} & \textbf{F1-RadGraph} &
  \textbf{Precision} & \textbf{Recall} & \textbf{F1-Score} &
  &  &  \\ 
\midrule
CheXagent & Test (OOD) & 
  3.00\,\textcolor{red}{$\downarrow$3.95} &
  13.50\,\textcolor{red}{$\downarrow$13.68} &
  46.00\,\textcolor{red}{$\downarrow$15.51} &
  4.50\,\textcolor{red}{$\downarrow$15.20} &
  30.50\,\textcolor{red}{$\downarrow$19.28} &
  40.00\,\textcolor{red}{$\downarrow$16.47} &
  33.00\,\textcolor{red}{$\downarrow$17.63} &
  &  &  \\
CheXpert-Plus & Test (OOD) &
  7.00\,\textcolor{red}{$\downarrow$7.84} &
  14.78\,\textcolor{red}{$\downarrow$13.23} &
  45.73\,\textcolor{red}{$\downarrow$15.03} &
  5.25\,\textcolor{red}{$\downarrow$16.89} &
  30.11\,\textcolor{red}{$\downarrow$18.63} &
  44.59\,\textcolor{red}{$\downarrow$3.01} &
  33.15\,\textcolor{red}{$\downarrow$13.33} &
  &  &  \\
MAIRA-2 & Test (OOD) &
  3.50\,\textcolor{red}{$\downarrow$4.62} &
  14.50\,\textcolor{red}{$\downarrow$13.32} &
  47.00\,\textcolor{red}{$\downarrow$15.30} &
  4.80\,\textcolor{red}{$\downarrow$15.57} &
  29.50\,\textcolor{red}{$\downarrow$19.22} &
  42.00\,\textcolor{red}{$\downarrow$15.91} &
  32.50\,\textcolor{red}{$\downarrow$17.86} &
  &  &  \\
\bottomrule
\end{tabular}%
}
\caption{CheXpert-Plus, CheXagent, and MAIRA-2 performance on the out-of-distribution HOPPR test set, showing deltas relative to their original Test results from Tables  \ref{tab:model_scores_main_impression} and \ref{tab:model_scores_main_findings}.}
\label{tab:hoppr_srrg_results_annotated_extended}
\end{table*}

\section{Related Work}

\textbf{Structured Reporting} \quad Chest X-ray reporting has long been characterized by a free-text narrative style, which, while flexible, can lack clarity and consistency~\cite{weiss2008structured, bosmans2012structured}. The lack of widespread standardization further reinforces this approach, as structured reporting templates, such as RSNA’s RadLex or BI-RADS for breast imaging, have not been universally adopted for CXRs. Studies have shown that even though structured reporting can improve completeness and diagnostic clarity~\cite{schwartz2011structured, bosmans2012structured}, many radiologists perceive it as rigid and less efficient compared to narrative reporting~\cite{bosmans2015structured}. Consequently, structured reporting remains underutilized, in part because CXRs require simultaneous assessment of multiple structures in context rather than in isolation~\cite{langlotz2002automatic}. \\

Given these challenges, efforts to standardize CXR reporting continue to face resistance, balancing the need for consistency with the flexibility required for nuanced clinical communication~\cite{dunnick2008report,kahn2009toward}. For systems aiming to generate automated or semi-automated reports from medical images, addressing this variability is crucial. Recent works in natural language processing and computer vision have attempted to handle the complexity of unstructured radiology reports, either by adopting standardized label sets derived from clinical knowledge bases or by using large-scale language models to learn patterns in free-text narratives. However, the gap between free-form clinical practice and structured data requirements remains a major challenge in achieving both clinical relevance and interoperability. \\

\textbf{Automated Radiology Reporting} \quad Prior work in radiology report generation has explored architectural innovations, reinforcement learning, and retrieval-based approaches. Architectural novelties include memory-driven transformers to retain key generation details~\cite{chen2020generating}, cross-modal memory networks to align images and text~\cite{chen-etal-2021-cross-modal}, and models incorporating prior medical knowledge graphs for structured report generation~\cite{liu2021exploring, liu2021auto}. Reinforcement learning has also been used to optimize factual correctness~\cite{pmlr-v106-liu19a, miura2021improving, delbrouck-etal-2022-improving}. Recently, larger models have been employed for radiology report generation. Notable examples include RaDialog~\cite{pellegrini2023radialog}, which integrates visual features and structured pathology findings with an LLM through parameter-efficient fine-tuning, and RGRG~\cite{tanida2023interactive}, a region-guided model that detects and describes anatomical regions to enhance transparency, interactivity, and explainability. Additionally, "LLM-sized" models such as MAIRA-2~\cite{bannur2024maira}, CheXagent~\cite{chen2024chexagent}, and MedVersa~\cite{zhou2024generalist} have also been introduced to further advance the field.

\section{Limitations} \label{sec:limitations}

Despite the promising results of our proposed Structured Radiology Report Generation (SRRG) framework, several limitations remain:

\paragraph{Synthetic Dataset \& Annotations} Our SRRG dataset was produced by reformulating free-form radiology reports into a structured format using LLMs. Although our methodology enforces strict desiderata to avoid hallucinations and preserve factual content, it remains challenging to verify all generated samples at scale. To mitigate inaccuracies, we conducted a comprehensive reader study involving five board-certified radiologists, as described in Appendix~\ref{sec:reader_study}. Nevertheless, the possibility of subtle inconsistencies or biases introduced by the LLMs cannot be fully excluded.

\paragraph{Fine-tuning Approaches} The range of model sizes and different training strategies used in our experiments (e.g., LoRA-based parameter-efficient fine-tuning for large models such as MAIRA-2 vs.\ full fine-tuning for smaller models) may affect the comparability of results. While these choices were made to accommodate computational feasibility, a standardized fine-tuning scheme across all models might yield a more uniform assessment of performance and could be explored in future work.

\paragraph{Reader Study Constraints} Our reader study focused on validating both structured reports and fine-grained disease labels derived from the SRR-BERT model. Although board-certified radiologists reviewed a representative sample of utterances, they occasionally encountered ambiguous cases where the available clinical context did not suffice to differentiate among closely related conditions (e.g., pneumonia, atelectasis, or aspiration). Additionally, rare findings not covered by our disease taxonomy were annotated under an \textit{“Other”} category, potentially oversimplifying certain nuanced clinical observations. Expanding the taxonomy or incorporating additional clinical context (e.g., lab values or clinical notes) may address these ambiguities in future iterations.

\paragraph{F1-SRR-BERT vs.\ F1-CheXbert} Directly comparing F1-Scores of SRR-BERT (with 55 disease labels) and CheXbert (with 14 labels) remains inherently imperfect due to the many-to-many relationship in label mapping. A single CheXbert class can correspond to multiple labels in our hierarchical disease ontology, and vice versa. Although we attempted a best-effort alignment, the lack of a one-to-one mapping between the label spaces makes straightforward performance comparisons challenging. Future work could improve this alignment by exploring probabilistic approaches or expert-guided hierarchical restructuring to reconcile label disparities.

\section{Contributions}
JBD led the project, curated the datasets, supervised the experimental workflow, and was the primary author of the manuscript. JX conducted the experiments related to the Disease Classification Models. JM, AT, and ZC were responsible for fine-tuning the models on the SRRG dataset. SO and AA contributed through early-stage brainstorming. KZI, AJ, ER, CB and MM served as reviewing radiologists, providing expert evaluation of the results. MV and CL offered guidance during the project.

\section*{Acknowledgements}
This work was supported in part by the Medical Imaging and Data Resource Center (MIDRC), funded by the National Institute of Biomedical Imaging and Bioengineering (NIBIB) of the National Institutes of Health under contract 75N92020D00021 and through The Advanced Research Projects Agency for Health (ARPA-H).
\bibliography{custom}

\appendix
\clearpage

\section{Potential Risks}
All experiments in this study are conducted using publicly available chest X-ray datasets (MIMIC-CXR and CheXpert Plus) that are fully deidentified, thereby minimizing risks related to patient privacy and data confidentiality. The text restructuring and disease label generation steps use GPT-4 deployed via Azure services, with the account explicitly configured to opt out of human data review.\\

While we believe releasing our models and code is valuable for advancing research, we emphasize that these models are for investigational and educational purposes only. They have not received regulatory approval for clinical deployment, and medical professionals must retain ultimate responsibility for diagnosis and patient management. As with all machine learning models, there is an inherent risk of errors or hallucinations, and predictions should be verified by qualified clinicians. We strongly encourage the community to apply robust validation, audits, and clinical oversight when exploring or extending our work.

\section{Reader Study} \label{sec:reader_study}

The reader study has been carried out by five board-certified radiologists from our institution on the annotation platform detailed in Appendix~\ref{app:reader_study_plat}. The following  examples and statistics summarize the textual changes between the original and edited impression sections. For each report pair, differences were quantified by counting word-level insertions, deletions, and replacements. The \textbf{similarity ratio} was computed using Python’s \texttt{difflib.SequenceMatcher} via 
\[
\resizebox{\columnwidth}{!}{$\displaystyle \text{Similarity Ratio} = \frac{2 \times \text{Matches}}{\text{Total Tokens in Original and Edited}}$}
\]
yielding a value between 0 (completely different) and 1 (identical).

\subsection*{Example 1: \texttt{mimic-53235571}}
\begin{lstlisting}[basicstyle=\small\ttfamily, breaklines=true, linewidth=\columnwidth, numbers=none]
Original Impression:
1. Bibasilar opacities that may be related to atelectasis, with a differential 
   including underlying infection, pneumonia, or aspiration.
2. New opacity in the lateral left mid lung, nonspecific but potentially 
   representing additional consolidation or pulmonary infarct.

Edited Impression:
1. Bibasilar opacities may be related to atelectasis, although underlying 
   infection, pneumonia, and/or aspiration is of concern.
2. New opacity in the lateral left mid lung, nonspecific but potentially 
   representing additional consolidation or pulmonary infarct.

Diff Stats:
Insertions: 0, Deletions: 1, Replacements: 9, Similarity Ratio: 0.82
\end{lstlisting}

\subsection*{Example 2: \texttt{mimic-59654440}}
\begin{lstlisting}[basicstyle=\small\ttfamily, breaklines=true, linewidth=\columnwidth, numbers=none]
Original Impression:
1. Resolving consolidation at the right lung base, likely due to dependent 
   edema or combined dependent edema and atelectasis.
2. Mild to moderate enlargement of the heart.
3. No pneumothorax.
4. Dual-channel dialysis catheter in situ with the tip in the right atrium.

Edited Impression:
1. Resolving consolidation at the right lung base with minimal residual 
   interstitial edema.

Diff Stats:
Insertions: 0, Deletions: 0, Replacements: 35, Similarity Ratio: 0.29
\end{lstlisting}

\subsection*{Impression Statistics}
\begin{lstlisting}[basicstyle=\small\ttfamily, breaklines=true, linewidth=\columnwidth, numbers=none]
Total studies reviewed: 233
Studies with changes: 130 (55.79%)
Average insertions per study: 0.42
Average deletions per study: 4.16
Average replacements per study: 4.50
Average similarity ratio: 0.77
\end{lstlisting}

Although 55.79\% of the impression exhibited changes, many modifications are subtly reflected by a relatively high overall similarity ratio. However, some reports demonstrate significant edits, underlining the need for enhanced clarity and precise clinical communication in the impression sections of CXR reports.

\subsection*{Findings Statistics}

\begin{lstlisting}[basicstyle=\small\ttfamily, breaklines=true, linewidth=\columnwidth,numbers=none]
Total studies reviewed: 233
Studies with changes: 164 (70.39%)
Average insertions per study: 4.97
Average deletions per study: 3.46
Average replacements per study: 4.64
Average similarity ratio: 0.88
\end{lstlisting}

The analysis reveals that a significant portion of the studies (70.39\%) underwent modifications, indicating that changes were applied in the majority of the cases. However, the higher average similarity ratio of 0.88 may suggest that these edits are relatively minor. On average, the modifications involved about 4.97 insertions, 3.46 deletions, and 4.64 replacements per study, which implies that while the impression sections were updated, the overall content remains largely consistent with the original. This balance indicates that the editing process likely focused on refining clarity and precision without altering the fundamental diagnostic information conveyed in the reports.

\subsection*{Utterance Label Consistency}

In this experiment, we assess the consistency of utterance labels extracted from the GPT models and compare them with manually reviewed labels. Two metrics are computed:
\begin{enumerate}
    \item \textbf{Exact match} GPT's labels and reviewed labels are the same.
    \item \textbf{Jaccard Similarity:} The ratio of the size of the intersection to the size of the union of the GPT's and reviewed label sets.
\end{enumerate}

The overall statistics from the evaluation are as follows:
\begin{lstlisting}[basicstyle=\small\ttfamily, breaklines=true, linewidth=\columnwidth, numbers=none]
Total utterances reviewed: 1609
Matched utterances: 1339
Exact Match Rate: 0.72
Average Jaccard Similarity: 0.74
\end{lstlisting}

These results indicate that, on average, 72\% of the consensus labels are present in the reviewed labels, and there is a 74\% overlap between the two label sets. The high similarity metrics suggest that the consensus approach is effective for capturing the expected labels across different sources, thereby validating our methodology for robust label extraction in utterances.

\section{Model Sizes and Hyperparameters}
MAIRA-2 uses an 87M-parameter ViT model, with its language model initialized from Vicuna 7B v1.5. We evaluated the 3B version of CheXagent-2. CheXpert-Plus is a SwinV2-based model with a BERT decoder (2 layers), while RaDialoG is a 7B-parameter model. For fine-tuning SRRG, we trained all the weights of CheXpert-Plus and CheXagent, using the default LoRA parameters from the Hugging Face PEFT library.

\section{Dataset Breakdown of Diseases} \label{app:disease_breakdown}

\begin{table}[!h]
    \centering
    \renewcommand{\arraystretch}{1.2}   
    \setlength{\tabcolsep}{6pt}         
    \caption{Dataset Breakdown for Upper Labels}
    \label{tab:upper_datasets}
    \resizebox{\columnwidth}{!}{%
    \begin{tabular}{lcc}
        \toprule
        \rowcolor{headerblue}
        \textbf{Anatomical Header / Category} & \textbf{Upper Levels} & \textbf{Num. Examples} \\ 
        \midrule
        \multirow{9}{*}{Lungs and Airways} 
            & \cellcolor{gray!10}Consolidation                     & \cellcolor{gray!10}340,867    \\ 
            & Diffuse air space opacity                            & 100,154    \\ 
            & \cellcolor{gray!10}Lung Finding                      & \cellcolor{gray!10}95,122     \\ 
            & Air space opacity                                    & 47,921     \\ 
            & \cellcolor{gray!10}Solitary masslike opacity         & \cellcolor{gray!10}40,831     \\ 
            & Focal air space opacity                              & 14,222     \\ 
            & \cellcolor{gray!10}Segmental collapse                & \cellcolor{gray!10}10,685     \\ 
            & Multiple masslike opacities                          & 547        \\ 
        \cmidrule(lr){2-3}
            & \cellcolor{gray!10}\textbf{Total}                    & \cellcolor{gray!10}650,349    \\ 
        \midrule
        \multirow{5}{*}{Pleura} 
            & Pleural Effusion                                     & 173,883    \\ 
            & \cellcolor{gray!10}Pneumothorax                       & \cellcolor{gray!10}56,706     \\ 
            & Pleural Thickening                                   & 31,210     \\ 
            & \cellcolor{gray!10}Pleural finding                    & \cellcolor{gray!10}7,734      \\ 
        \cmidrule(lr){2-3}
            & \cellcolor{gray!10}\textbf{Total}                    & \cellcolor{gray!10}269,533    \\ 
        \midrule
        \multirow{3}{*}{Cardiovascular} 
            & \cellcolor{gray!10}Widened cardiac silhouette         & \cellcolor{gray!10}58,189     \\ 
            & Vascular finding                                     & 20,480     \\ 
        \cmidrule(lr){2-3}
            & \cellcolor{gray!10}\textbf{Total}                    & \cellcolor{gray!10}78,669     \\ 
        \midrule
        \multirow{4}{*}{Hila and Mediastinum} 
            & Widened aortic contour                               & 17,513     \\ 
            & \cellcolor{gray!10}Mediastinal finding                & \cellcolor{gray!10}13,779     \\ 
            & Mediastinal mass                                     & 5,922      \\ 
        \cmidrule(lr){2-3}
            & \cellcolor{gray!10}\textbf{Total}                    & \cellcolor{gray!10}37,214     \\ 
        \midrule
        \multirow{4}{*}{Musculoskeletal and Chest Wall} 
            & \cellcolor{gray!10}Fracture                           & \cellcolor{gray!10}34,192     \\ 
            & Chest wall finding                                   & 11,614     \\ 
            & \cellcolor{gray!10}Musculoskeletal finding            & \cellcolor{gray!10}617        \\ 
        \cmidrule(lr){2-3}
            & \cellcolor{gray!10}\textbf{Total}                    & \cellcolor{gray!10}46,423     \\ 
        \midrule
        Abdominal       & \cellcolor{gray!10}Subdiaphragmatic gas               & \cellcolor{gray!10}3,475      \\ 
        \midrule
        Support Devices & Support Devices                                      & 96,274     \\ 
        \midrule
        No Finding      & \cellcolor{gray!10}–                                  & \cellcolor{gray!10}600,328    \\ 
        \bottomrule
    \end{tabular}
    }
\end{table}

\clearpage
\begin{figure*}[h]
  \centering
  \captionsetup{labelformat=empty}
  \caption{}          
  \label{app:struct_prompt}
  \begin{adjustbox}{max height=0.95\textheight}
  
\begin{tcolorbox}[title=\textit{Structuring Prompt}, colframe=black!75!white, colback=white, sharp corners]
Your task is to improve the formatting of a radiology report, ensuring it is \textbf{clear, concise, and well-structured} with appropriate section headings.

\textbf{Guidelines:}
\begin{enumerate}
    \item \textbf{Section Headers:} Each section should begin with a section header followed by a colon. Include only the relevant information as specified.
    \item \textbf{Identifiers:} Remove any sentences containing identifiers such as dates, surnames, first names, healthcare providers, vendors, or institutions. \textbf{Important:} Retain sex and age information if present.
    \item \textbf{Findings and Impression Sections:} Focus exclusively on the \textbf{current examination results}. Do not reference previous studies or historical data.
    \item \textbf{Content Restrictions:} Strictly include only content relevant to the structured sections provided. Do not add or extrapolate beyond the original report.
\end{enumerate}

\textbf{Sections to Include (if applicable):}
\begin{enumerate}
    \item \textbf{Exam Type:} Specify the type of examination conducted.
    \item \textbf{History:} Provide a brief clinical history and state the clinical question or suspicion prompting the imaging.
    \item \textbf{Technique:} Describe the examination technique and any specific protocols used.
    \item \textbf{Comparison:} Indicate prior imaging studies reviewed for comparison.
    \item \textbf{Findings:} List all positive and relevant negative observations for each organ system under structured headers.
\end{enumerate}

\textbf{Template for Findings:}
\begin{verbatim}
Header 1:
- Observation 1
- ...
Header 2:
- Observation 1
- Observation 2
- ...
...
\end{verbatim}

\textbf{Use only the following headers for organ systems:}
\begin{itemize}
    \item Lungs and Airways
    \item Pleura
    \item Cardiovascular
    \item Hila and Mediastinum
    \item Tubes, Catheters, and Support Devices
    \item Musculoskeletal and Chest Wall
    \item Abdominal
    \item Other
\end{itemize}

\textbf{Important:} \textit{Do not use any headers other than those listed above. Only use the specified headers corresponding to the organ systems mentioned in the original radiology report.}

\textbf{6. Impression:} Summarize the key findings in a numbered list, ranking them from most to least clinically relevant.

\textbf{The radiology report to improve is the following:}
\begin{verbatim}
{}
\end{verbatim}

\end{tcolorbox}
\end{adjustbox}
\end{figure*}

\begin{figure*}[h]
  \centering
  \captionsetup{labelformat=empty}
  \caption{}          
  \label{app:disease_prompt}
  \begin{adjustbox}{max height=0.95\textheight}
\begin{tcolorbox}[title=\textit{Diseases prompt}, colframe=black!75!white, colback=white, sharp corners]

Your task is to identify the diseases discussed in chest X-ray findings. You will be provided with:  
\textbf{1) Instructions}  \\
\textbf{2) A list of possible diseases}    \\
\textbf{3) A list of chest X-ray findings}    \\

\textbf{1) Instructions:}  
Your task is to provide the following:  
\begin{enumerate}[label=\alph*)]
    \item The diseases that are present as a numbered list. There can be zero, one, or multiple diseases discussed. If no disease is present or discussed in a finding, answer: \texttt{"1. No Finding"} for that finding.
    \item The status of the disease discussed. The status can be:
    \begin{itemize}
        \item \textbf{Present}: The disease is confirmed to be present in the patient.
        \item \textbf{Absent}: The disease is confirmed to be not present in the patient.
        \item \textbf{Uncertain}: It is unclear whether the disease is present or absent, often due to inconclusive test results or insufficient information.
    \end{itemize}
\end{enumerate}

Below is the template to provide your answer. You must respect this format and not provide any explanations or additional content:

\begin{verbatim}
<finding 1> => 1. <disease 1> (Present) 2. <disease 2> (Uncertain)
<finding 2> => 1. <disease 1> (Absent)
...
\end{verbatim}

\textbf{2) List of possible diseases:}  
\begin{itemize}
    \item No Finding
    \item Lung Lesion
    \item Edema
    \item Pneumonia
    \item Atelectasis
    \item Lung collapse
    \item Perihilar airspace opacity
    \item Air space opacity–multifocal
    \item Mass/Solitary lung mass
    \item Nodule/Solitary lung nodule
    \item Cavitating mass with content
    \item Cavitating masses
    \item Emphysema
    \item [...]
\end{itemize}

\textbf{3) List of chest X-ray findings (one per line):}  
\begin{verbatim}
{}
\end{verbatim}

\end{tcolorbox}
\end{adjustbox}
\end{figure*}

\begin{figure*}[t] 
  \centering
  \captionsetup{labelformat=empty}
  \caption{}          
  \label{app:disease tree}
  \begin{adjustbox}{max height=0.95\textheight}
\begin{tcolorbox}[title=\textit{Diseases Tree}, colframe=black!75!white, colback=white, sharp corners]
\begin{verbatim}
1. No Finding
2. Lung Finding
   2.1. Lung Opacity
       2.1.1. Air space opacity
           2.1.1.1. Diffuse air space opacity
               2.1.1.1.1. Edema
           2.1.1.2. Focal air space opacity
               2.1.1.2.1. Consolidation
                   2.1.1.2.1.1. Pneumonia
                   2.1.1.2.1.2. Atelectasis
                   2.1.1.2.1.3. Aspiration
               2.1.1.2.2. Segmental collapse
                   2.1.1.2.2.1. Lung collapse
               2.1.1.2.3. Perihilar airspace opacity
           2.1.1.3. Air space opacity–multifocal
       2.1.2. Masslike opacity
           2.1.2.1. Solitary masslike opacity
               2.1.2.1.1. Mass/Solitary lung mass
               2.1.2.1.2. Nodule/Solitary lung nodule
               2.1.2.1.3. Cavitating mass with content
           2.1.2.2. Multiple masslike opacities
               2.1.2.2.1. Cavitating masses
   2.2. Emphysema
   2.3. Fibrosis
   2.4. Pulmonary congestion
   2.5. Hilar lymphadenopathy
   2.6. Bronchiectasis
3. Pleural Finding
   3.1. Pneumothorax
       3.1.1. Simple pneumothorax
       3.1.2. Loculated pneumothorax
       3.1.3. Tension pneumothorax
   3.2. Pleural Thickening
       3.2.1. Pleural Effusion
           3.2.1.1. Simple pleural effusion
           3.2.1.2. Loculated pleural effusion
       3.2.2. Pleural scarring
   3.3. Hydropneumothorax
   3.4. Pleural Other
4. Widened Cardiac Silhouette
   4.1. Cardiomegaly
   4.2. Pericardial effusion
5. Mediastinal Finding
   5.1. Mediastinal Mass
       5.1.1. Inferior mediastinal mass
       5.1.2. Superior mediastinal mass
   5.2. Vascular Finding
       5.2.1. Widened aortic contour
           5.2.1.1. Tortuous Aorta
       5.2.2. Calcification of the Aorta
       5.2.3. Enlarged pulmonary artery
   5.3. Hernia
   5.4. Pneumomediastinum
   5.5. Tracheal deviation
6. Musculoskeletal Finding
   6.1. Fracture
       6.1.1. Acute humerus fracture
       6.1.2. Acute rib fracture
       6.1.3. Acute clavicle fracture
       6.1.4. Acute scapula fracture
       6.1.5. Compression fracture
   6.2. Shoulder dislocation
   6.3. Chest wall finding
       6.3.1. Subcutaneous Emphysema
7. Support Devices
   7.1. Suboptimal central line
   7.2. Suboptimal endotracheal tube
   7.3. Suboptimal nasogastric tube
   7.4. Suboptimal pulmonary arterial catheter
   7.5. Pleural tube
   7.6. PICC line
   7.7. Port catheter
   7.8. Pacemaker
   7.9. Implantable defibrillator
   7.10. LVAD
   7.11. Intraaortic balloon pump
8. Upper Abdominal Finding
   8.1. Subdiaphragmatic gas
       8.1.1. Pneumoperitoneum
\end{verbatim}
\end{tcolorbox}
\end{adjustbox}
\twocolumn
\end{figure*}

\clearpage
\onecolumn
\section{Reader Study Platform} \label{app:reader_study_plat}

\begin{figure}[!htbp]
    \centering
    \includegraphics[width=\textwidth]{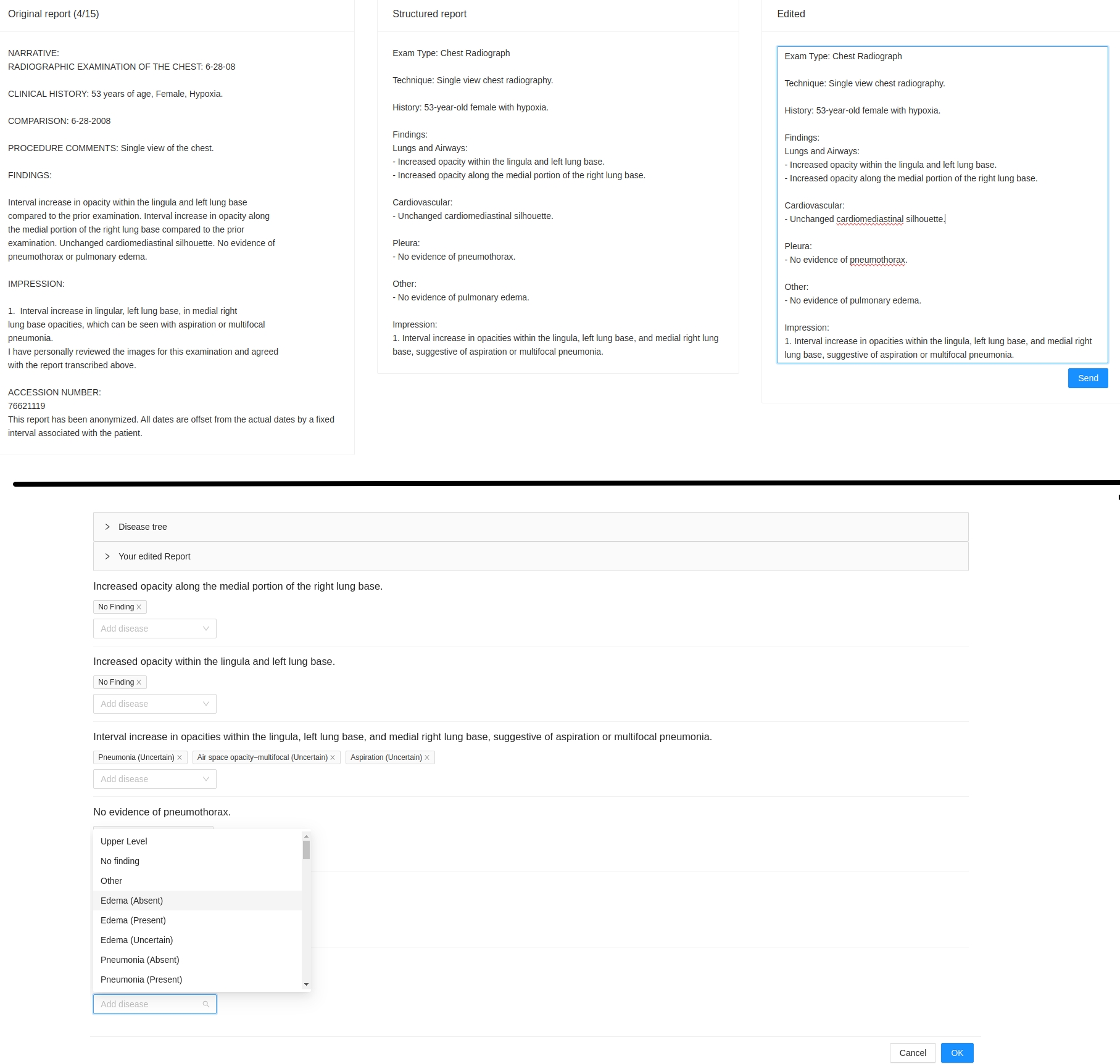}
    \caption{This figure illustrates our reader study annotation workflow. At the top, the radiologist sees the original report (left), the GPT-generated structured report (middle), and an editable text box (right). At the bottom, after validating the structured report, the radiologist annotates each utterance. The labels for these utterances are pre-filled based on the GPT model’s consensus. Throughout this process, the radiologist can consult both the edited report and a disease tree to guide the labeling.}
\end{figure}

\end{document}